\documentclass{article}

\usepackage{arxiv}

\usepackage[utf8]{inputenc} 
\usepackage[T1]{fontenc}    
\usepackage{hyperref}       
\usepackage{url}            
\usepackage{booktabs}       
\usepackage{amsfonts}       
\usepackage{nicefrac}       
\usepackage{microtype}      
\usepackage{lipsum}
\usepackage{graphicx}

\usepackage{amsmath}
\usepackage{algorithm}
\usepackage{algpseudocode}

\title{Spiffy: Efficient Implementation of CoLaNET for Raspberry Pi}

\author{
Andrey Derzhavin \\
  Chuvash State University, \\
  Cheboksary, Russia\\
\And
Denis Larionov \\
  Chuvash State University, \\
  Cheboksary, Russia\\
  Cifrum, Moscow, Russia\\
}

\begin{document}
\maketitle
\begin{abstract}
This paper presents a lightweight software-based approach for running spiking neural networks (SNNs) without relying on specialized neuromorphic hardware or frameworks. Instead, we implement a specific SNN architecture (CoLaNET) in Rust and optimize it for common computing platforms. As a case study, we demonstrate our implementation, called Spiffy, on a Raspberry Pi using the MNIST dataset. Spiffy achieves 92\% accuracy with low latency - just 0.9 ms per training step and 0.45 ms per inference step. The code is open-source.
\end{abstract}


\section{Introduction}

According to \cite{Ivanov2022}, modern artificial intelligence (AI) systems still fall far short of their biological counterparts in terms of energy efficiency, adaptability, scalability, and versatility. A promising approach to bridging this gap lies in neuromorphic computing — the emulation of the brain’s organizational and functional principles.

A key distinction between biological brains and artificial neural networks is the spiking nature of information transmission. Spiking neural networks (SNNs) \cite{Gerstner2014, Kiselev2020a} model this communication paradigm, offering neuromorphic advantages such as event-driven (asynchronous) processing, local learning rules, and sparse computations \cite{Ivanov2022}.

Historically, AI breakthroughs have been tightly coupled with advances in hardware \cite{Hooker2020}. Thus, co-designing algorithms and hardware is critical. SNNs enable efficient hardware implementations by replacing dense matrix multiplications with additive operations (and rare multiplications). Unlike gradient-based methods \cite{Rumelhart1986}, local learning rules eliminate the need for global synchronization across compute nodes \cite{Moraitis2021}. Moreover, their inherently asynchronous computation model facilitates scalability.

A promising architecture that additionally incorporates neuromorphic ideas of columnar network organization, competition, and modulated  plasticity is the Columnar Layered Network (CoLaNET) architecture \cite{Kiselev2024a, Kiselev2024b, Kiselev2025b}. CoLaNET employs local learning rules and is designed for classification tasks (supervised learning). A distinctive feature of CoLaNET, besides its columnar organization, is the combination of anti-Hebbian plasticity, which leads to the degradation of weights, and modulated (dopamine) plasticity, which increases the weights, compensating for the degradation and enabling useful learning.

CoLaNET is developed and tested using a proprietary simulation framework called ArNI-X \cite{ArNI-X}. This framework helps researchers efficiently study SNN algorithms and architectures on both CPUs and GPUs. However, ArNI-X is not designed for real-world applications or performance optimization. Another issue is that ArNI-X is proprietary software, meaning other scientists can’t access the code to verify or replicate published results.

This paper presents the first hardware implementation of a SNN based on CoLaNET — called Spiffy. Spiffy retains the CoLaNET architecture but introduces key optimizations for both computational efficiency and accuracy (Sec. \ref{sec:architecture}). Specifically, it uses a simplified synaptic weight-resource function \cite{Kiselev2017} and slightly modifies the modulated plasticity mechanism. The Spiffy software is implemented in Rust, making it portable across multiple platforms (x86/x64/ARM/RISC-V, Android/iOS, Arduino/Raspberry Pi, WebAssembly/KVM). With further optimizations, Spiffy achieves 92\% accuracy on the MNIST dataset \cite{LeCun1998} using a single network instance — outperforming the original CoLaNET.

When running Spiffy on Raspberry Pi4 in training mode (on-device local learning), it demonstrates a latency of 0.9 ms per computation cycle during image processing, and 0.45 ms per cycle in inference mode. During silent periods (in both training and inference), the latency is reduced to 0.1 ms per cycle. The source code and experimental results are available as open-source at https://gitflic.ru/project/algidrus/spiffy.

The key contribution of this work is demonstrating a complete end-to-end pipeline for SNN training with local rules on edge devices without relying on proprietary tools or frameworks. All the CoLaNET algorithms are manually implemented in Rust, enabling Spiffy to run on common hardware architectures without requiring specialized neuroprocessors. The optimization approaches developed during Spiffy's implementation achieve high classification accuracy without using neural network ensembles, reducing the total network size by 15× compared to the original CoLaNET. Spiffy serves as a benchmark for developing SNN-based solutions on popular hardware platforms.

\section{Methods}
\label{sec:methods}

\subsection{CoLaNET on MNIST}
\label{sec:colanet}

The MNIST dataset \cite{LeCun1998} contains 70,000 images of handwritten digits, divided into 10 classes (60,000 for training and 10,000 for testing). Each image is single-channel and has a resolution of 28×28 pixels. The intensity value of each pixel is encoded as a number ranging from 0 to 255.

In \cite{Kiselev2025b}, a configuration of CoLaNET obtained through hyperparameter optimization using a genetic algorithm is presented. The network for MNIST classification consists of 15 separate CoLaNET networks. Each network contains 10 columns (corresponding to the number of classes), and each column comprises 15 microcolumns. The total number of neurons is 7,050, with 2,250 of them having plastic synapses. This configuration achieves 95\% accuracy when trained for a single epoch, while the individual networks in the ensemble do not exceed 75\% accuracy. An optimized single-network CoLaNET configuration demonstrates 89\% accuracy.

\subsection{Spiffy architecture}
\label{sec:architecture}

\begin{figure}[htbp]
  \centering
  \includegraphics[width=0.8\textwidth]{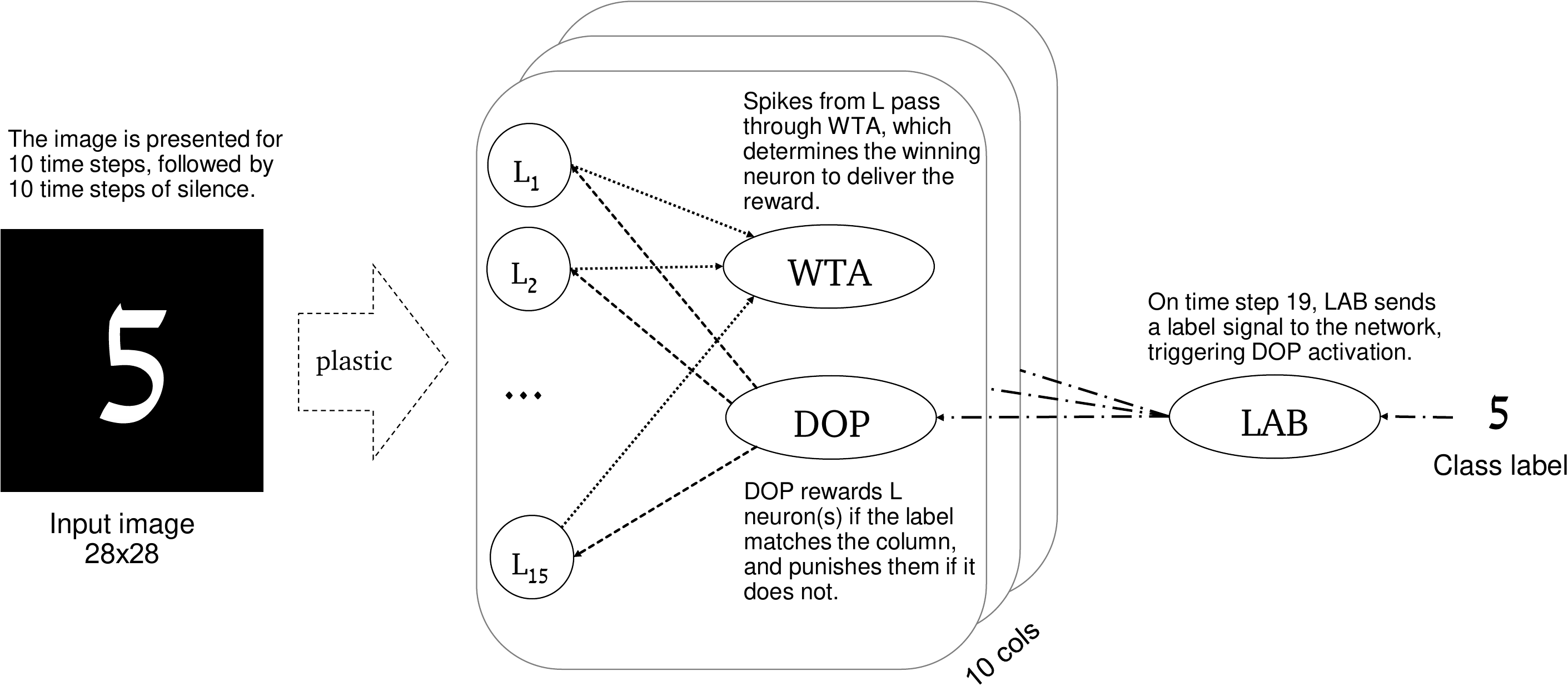}
  \caption{Spiffy - training}
  \label{fig:training}
\end{figure}

Spiffy is built on a single instance of the CoLaNET \cite{Kiselev2024a, Kiselev2024b, Kiselev2025b} and represents a fully connected SNN with a columnar structure. It contains a fixed number of columns and a single shared LAB module responsible for presenting labels during supervised learning. The number of columns corresponds to the number of classes in the classification task. Each column in Spiffy consists of 15 plastic neurons (L), a dopamine pump (DOP), and a winner-take-all (WTA) module (Fig. \ref{fig:training}). Unlike CoLaNET, in Spiffy, the modules responsible for modulated plasticity and competition (DOP and WTA) are not implemented as separate neurons but are instead realized programmatically for efficiency. Thus, Spiffy contains 150 plastic neurons (the same as a single CoLaNET network).

To convert pixel intensities into spike sequences, CoLaNET uses rate coding, where the probability of generating a spike at a given time step is proportional to the pixel intensity, with an intensity of 0 corresponding to zero probability and 255 to the maximum probability. Objects are presented to the network for 10 time steps, followed by 10 steps of silence. Spiffy also uses 10 time steps for presentation and 10 for silence, but unlike CoLaNET, the number of spikes over 10 steps is explicitly determined (proportional to intensity) and uniformly distributed across the entire interval. Given that pixel intensities do not change during the presentation of a single image, this approach is equivalent to CoLaNET's probabilistic coding, adjusted for the absence of stochasticity.

Stochasticity in Spiffy is achieved, in part, through random initialization of the weights of L neurons, whereas in CoLaNET, weights are initialized to zero.

\begin{figure}[htbp]
  \centering
  \includegraphics[width=0.5\textwidth]{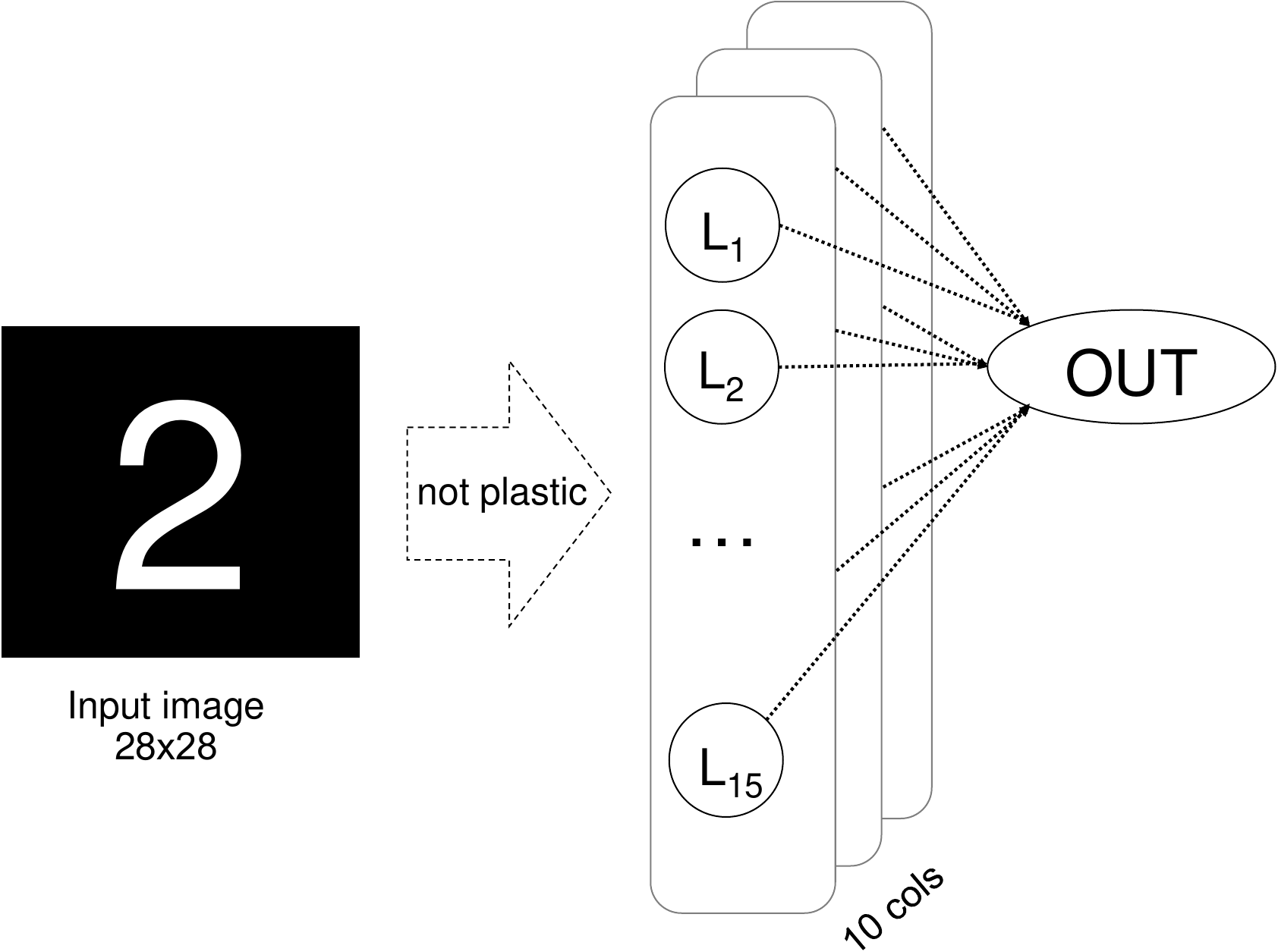}
  \caption{Spiffy - inference}
  \label{fig:inference}
\end{figure}

If the network is trained (L neurons are non-plastic), the class of the presented object is determined by the OUT module based on the maximum spike count from a column (Fig. \ref{fig:inference}). Unlike CoLaNET, Spiffy does not use the WTA mechanism during inference. Thus, all L neurons can potentially generate spikes, rather than only the winner (as in CoLaNET).

During the training phase, on the final time step of the corresponding interval (step 19), the label is presented via the LAB component, and the DOP is activated (Fig. \ref{fig:training}). The DOP governs plasticity mechanisms, balancing modulated plasticity with anti-Hebbian plasticity. However, unlike in CoLaNET, in Spiffy, modulated plasticity in some cases leads to an increase in the weights of all L neurons in a given column, rather than only the single L neuron that fired the first spike.

More formally, DOP operates in either a reward or punishment mode. Synaptic potentiation is always associated with reward and can be either single or group-based. A single reward occurs when an L neuron fires correctly in accordance with the presented label. If multiple L neurons in the same column fire simultaneously, the winner is determined by the WTA mechanism (at step 19). A group-based reward, applied to all L neurons in the column, occurs when none of the L neurons generate a spike (in the column corresponding to the correct label) during the image presentation period. A complete description of the plasticity algorithm is provided in Listing \ref{alg:spiffy}.

\begin{algorithm}
\caption{Spiffy's plasticity rule}
\label{alg:spiffy}
\begin{algorithmic}[1]
\For{each input image}
    \State reset correctly spiking neurons list \textbf{REW}
    \State reset incorrectly spiking neurons list \textbf{PUN}
    
    \For{10 presentation timesteps}        
        \For{each neuron $L$}
            \If{neuron fired for the correct label}
                \State add to \textbf{REW}
            \ElsIf{neuron fired wrongly}
                \State add to \textbf{PUN}
            \EndIf
        \EndFor
    \EndFor

    \For{10 silence timesteps}
        \If{timestep == 19}            
            \If{\textbf{REW} is empty}
                \State increase weights of all neurons in the correct label column by dopamine quantum (group update)
            \Else
                \State increase weight of a random neuron from \textbf{REW} by dopamine quantum (modulated plasticity)
            \EndIf
            
            \For{each neuron in \textbf{PUN}}
                \State decrease weight by dopamine quantum (anti-Hebbian plasticity)
            \EndFor
        \EndIf
    \EndFor    
\EndFor
\end{algorithmic}
\end{algorithm}

Synaptic depression is triggered by the punishment signal from the DOP and occurs when some of the L neurons in a given column fire incorrectly (i.e., during the presentation of an image from a different class). In this case, the synapses of such neurons are depressed at the 19th time step of the image presentation.

\subsection{Neuron model and plasticity}
\label{sec:plasticity}

Similar to CoLaNET, L neurons employ the leaky integrate and fire (LIF) neuron model along with a current-based delta synapse model. This means that whenever a synapse receives a spike, it instantaneously changes the membrane potential by the value of the synaptic weight $w$. Thus, the state of a neuron at time $t$ is described by its membrane potential $u(t)$, the dynamics of which are governed by the equation:

\begin{equation}
\frac{du}{dt} = -\frac{u}{\tau_v} + \sum_{i,k} w_i \delta(t - t_{ik}),
\label{eq:lif}
\end{equation}

\noindent with the condition that if $u$ exceeds 1, the neuron fires, and the value of $u$ is reduced by 1. The remaining notations in (\ref{eq:lif}) have the following meanings: $\tau_v$ is the membrane leak time constant; $w_i$ is the weight of the i-th synapse; and $t_{ik}$ is the time at which the i-th synapse received its k-th spike (among all spikes arriving at the neuron's synapses).

In both CoLaNET and Spiffy, plasticity rules are applied not to the synaptic weight itself but to an associated variable -- the synaptic resource $W$. This approach addresses the issue of uncontrolled weight growth. In CoLaNET, the dependence of the weight $w$ on the resource $W$ is defined by the following equation:

\begin{equation}
w = w_{\text{min}} + \frac{(w_{\text{max}} - w_{\text{min}}) \max(W,0)}{w_{\text{max}} - w_{\text{min}} + \max(W,0)} ,
\label{eq:syn_res}
\end{equation}

\noindent where $w_{\text{min}} \leq w < w_{\text{max}}$, $w$ monotonically increasing as a function of $W$ \cite{Kiselev2017}. In Spiffy, in addition to Eq. \ref{eq:syn_res}, a linear dependence is also investigated:

\begin{equation}
w = \min(w_{\text{max}}, \max(w_{\text{min}}, W)) ,
\label{eq:linear}
\end{equation}

\noindent where $w_{min}$ and $w_{max}$ represent the minimum and maximum values of the weight $w$. This direct dependence proves more efficient while not causing any significant degradation in classification accuracy on MNIST.

Certain original CoLaNET mechanisms are not implemented in the current Spiffy configuration. These include: adaptive threshold mechanism, synaptic resource renormalization, virtual synapses, and explicit control of the plasticity window under modulated learning conditions.

\subsection{Rust implementation}
\label{sec:rust}

The Spiffy network is implemented in the Rust programming language. Rust compiles to native machine code, ensuring compatibility with most modern platforms. The source code of Spiffy is open-source and available for download at https://gitflic.ru/project/algidrus/spiffy. The build process is detailed in the spiffy.md document located in the docs folder of the Spiffy project.

\section{Experiments}
\label{sec:experiments}

The reverse engineering of the CoLaNET architecture was implemented in several steps. First, using access to the ArNI-X source code, the basic modules such as the neuron model and synaptic resource functions were reproduced and covered with tests. The remaining mechanics, such as winner-take-all (WTA), Hebbian plasticity, modulated plasticity, and others, were implemented based on the conceptual representation of the CoLaNET architecture, so their software implementation differs significantly from ArNI-X. Moreover, ArNI-X currently employs certain limitations (e.g., 1024 synapses per neuron) related to efficient GPU-based implementation. Spiffy does not have such restrictions.

The initial implementation of Spiffy, which evolved through iterative refinement of CoLaNET mechanics, showed relatively low accuracy of around 76\% (compared to 89\% for CoLaNET). An effective solution to achieve better metrics than CoLaNET for a single network turned out to be a minor adjustment of CoLaNET's core mechanics. Specifically, this involved using a linear weight-to-synaptic-resource dependency (Eq. \ref{eq:linear}), random weight initialization, and an additional group update mechanism (Sec. \ref{sec:architecture}).

The linear resource dependency function (Eq. \ref{eq:linear}) is computationally more efficient, random weight initialization introduces stochasticity and is a more natural form of initialization for continuous learning, and the group update approach (Sec. \ref{sec:architecture}) simplifies modulated learning mechanics (compared to CoLaNET). However, it also leads to negative consequences, such as strong similarity between classified patterns among L neurons within a single column (as will be shown later).

Table \ref{tab:linear} presents the results of the best Spiffy configuration, which demonstrates classification accuracy of 91.08\% over 10 independent experiments on the MNIST dataset. The best configuration was obtained by optimizing Spiffy’s hyperparameters using the AdamW method. The optimization was performed on Intel I9-14900k (192 GB RAM) and AMD Ryzen9 9950x3D (128 GB RAM) machines. The columns in Table \ref{tab:linear} show the overall classification accuracy and per-class accuracy. The network performs best on digits 0 and 1, achieving 98\%, and worst on digit 9 (84\%).

\begin{table}[ht]
\centering
\caption{Spiffy's classification accuracy of the MNIST dataset. The configuration uses random weight initialization and linear synaptic resource function. Columns correspond to the MNIST classes.}
\label{tab:linear}
\begin{tabular}{|l|c|c|c|c|c|c|c|c|c|c|c|}
\hline
 & avg & 0 & 1 & 2 & 3 & 4 & 5 & 6 & 7 & 8 & 9 \\
\hline
acc & \textbf{91.08} & 97.92 & 98.63 & 88.50 & 89.34 & 92.52 & 87.59 & 95.16 & 92.23 & 84.72 &  84.21 \\
\hline
std & \textbf{0.26} & 0.48 & 0.26 & 2.42 & 2.68 & 1.66 & 2.72 & 1.08 & 1.11 & 2.88 & 3.50  \\
\hline
\end{tabular}
\end{table}

The experimental results with the classical synaptic resource function (Eq. \ref{eq:syn_res}) are presented in Table \ref{tab:classic}. The classification accuracy after optimization tested on 10 independent runs reaches 89.84\%, which is slightly lower than the accuracy achieved with the linear synaptic resource function (Eq. \ref{eq:linear}) — despite the linear function being more computationally efficient.

\begin{table}[ht]
\centering
\caption{The configuration uses random weight initialization and classic synaptic resource function.}
\label{tab:classic}
\footnotesize
\begin{tabular}{|l|c|c|c|c|c|c|c|c|c|c|c|}
\hline
& avg & 0 & 1 & 2 & 3 & 4 & 5 & 6 & 7 & 8 & 9 \\
\hline
acc & \textbf{89.84} & 97.23 & 98.10 & 89.42 & 89.39 & 92.66 & 85.33 & 94.78 & 91.72 & 78.13 & 81.64 \\
\hline
std & \textbf{0.51} & 0.90 & 0.45 & 1.65 & 2.04 & 1.88 & 3.37 & 1.16 & 0.95 & 4.99 & 3.38 \\
\hline
\end{tabular}
\end{table}

The impact of weight initialization was tested in an experiment using a linear synaptic resource function (Eq. \ref{eq:linear}) and zero-initialized weights. The results are presented in Table \ref{tab:zero}. The classification accuracy  after optimization tested on 10 independent runs reaches 89.87\%, which is slightly lower than that with random initialization. A key feature of the zero-initialization approach is that the heatmaps of the learned L-neuron receptive fields are more representative (as they contain no noise).

\begin{table}[ht]
\centering
\caption{The configuration uses zero weight initialization and linear synaptic resource function.}
\label{tab:zero}
\footnotesize
\begin{tabular}{|l|c|c|c|c|c|c|c|c|c|c|c|}
\hline
& avg & 0 & 1 & 2 & 3 & 4 & 5 & 6 & 7 & 8 & 9 \\
\hline
acc & \textbf{89.87} & 98.37 & 98.50 & 88.46 & 85.18 & 92.57 & 87.63 & 94.09 & 93.80 & 87.08 & 73.04 \\
\hline
std & \textbf{0.35} & 0.17 & 0.08 & 0.67 & 1.15 & 0.69 & 1.14 & 0.47 & 0.49 & 1.20 & 3.08 \\
\hline
\end{tabular}
\end{table}

\begin{figure}[ht]
  \centering
  \includegraphics[width=1\textwidth]{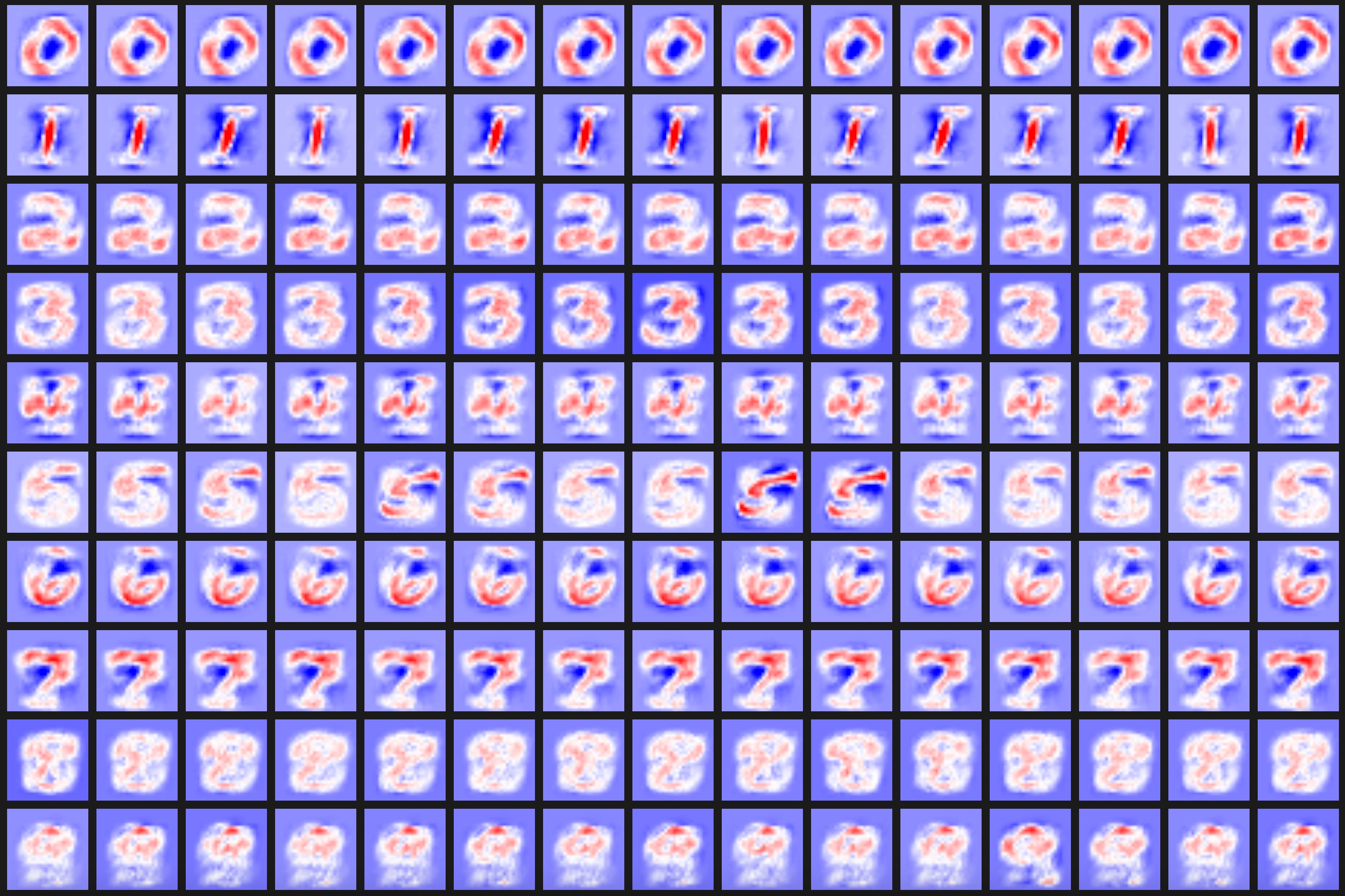}
  \caption{Heatmaps of Spiffy network L-layer neuron receptive fields post-training on MNIST. Weights are zero-initialized. Each row represents a digit class (0–9). Blue indicates negative weights; red indicates positive weights.}
  \label{fig:receptors}
\end{figure}

Fig. \ref{fig:receptors} shows the receptive fields of L neurons in the Spiffy network after training on MNIST (with weights initialized to zero). Each row corresponds to one column and contains different class representations (15 in total, matching the number of L neurons). The figure reveals that for classes 0 and 1, where Spiffy achieves the highest accuracy, the heatmaps exhibit high contrast, with class patterns clearly distinguishable. However, it can also be observed that the class prototypes strongly resemble each other, which is a consequence of group updates (and a drawback compared to CoLaNET). Mitigating this effect is a direction for future research.

In the final stage of the experiments, the Spiffy network was launched in training mode (on-device local learning) on a Raspberry Pi4 device with the Raspberry Pi OS (Debian 12 x64) operating system. The process of compiling Spiffy for Raspberry Pi4 does not differ from the compilation process for any other platform supporting the Rust compiler and the standard $std$ library. The Spiffy source code includes built-in debugging mechanisms that allow precise timestamping of individual events with high accuracy. When running Spiffy on Raspberry Pi4 in training mode, the observed latency is 0.9 ms per computation cycle (1 timestep in Alg. \ref{alg:spiffy}) when processing an input image and 0.45 ms per cycle in inference mode. During silence periods (both in training and inference), the latency is 0.1 ms per cycle.

\section{Conclusion}
\label{sec:conclusion}

The CoLaNET architecture represents a current research focus in SNNs. Various research groups are investigating CoLaNET implementations on FPGAs, the AltAI processor \cite{AltAI}, and within the Kaspersky Neuromorphic Platform (KNP) framework \cite{KNP}. This work presents the first hardware implementation of the CoLaNET architecture, which can serve as a benchmark for comparison with future neuromorphic AI systems.

Developing hardware systems with spiking neural networks doesn't necessarily require using SNN simulation frameworks followed by network placement onto neuroprocessors or FPGAs. An alternative approach involves direct software implementation of a specific SNN architecture in Rust, which can then be ported to various popular platforms. Given the widespread availability of these platforms, this method enables straightforward SNN adoption for AI system development. 

While this approach may potentially yield lower performance compared to neuroprocessor implementations, it offers significant advantages: implementation simplicity and speed, greater flexibility, unlimited functionality (for example, on-device learning capabilities), broad compatibility with mainstream platforms, and complete independence from proprietary tools.

Through Spiffy's algorithmic optimizations and support for additional optimization techniques - including pruning, quantization, and feature hierarchy implementation - CoLaNET-based solutions can achieve significantly improved hardware efficiency.

\bibliographystyle{unsrt}
\bibliography{ref}

\end{document}